# Contextualising Levels of Language Resourcedness that affect NLP tasks

***C. Maria Keet[1,2] and Langa Khumalo[3]***

*[1]Department of Computer Science, University of Cape Town, South Africa*

*[2]Computer Science Division, Stellenbosch University, South Africa*

*[3]South African Digital Language Resources Centre, South Africa*

**Abstract**

Several widely used software applications involve some form of processing of natural language, with tasks ranging from digitising hardcopies and text processing to speech generation. Varied language resources are used to develop software systems to accomplish a wide range of natural language processing (NLP) tasks, such as the ubiquitous spellcheckers and chatbots. Languages are typically characterised as either low (LRL) or high resourced languages (HRL) with African languages having been characterised as resource-scarce languages and English by far the most well-resourced language. But what lies in-between? We argue that the dichotomous typology of LRL and HRL for all languages is problematic. Through a clear understanding of language resources situated in a society, a matrix is developed that characterises languages as Very LRL, LRL, RL, HRL and Very HRL. The characterisation is based on the typology of *contextual features* for each category, rather than counting tools. The motivation is provided for each feature and each characterisation. The contextualisation of resourcedness, with a focus on African languages in this paper, and an increased understanding of where on the scale the language used in a project is, may assist in, among others, better planning of research and implementation projects. We thus argue in this paper that the characterisation of language resources within a given scale in a project is an indispensable component, particularly for those in the lower half of the scale.

**Short abstract:** A language resourcedness classification scheme based on a typology of contextual features, focussing on low-resourced languages in Africa.

## 1. Introduction

Natural language-intensive projects, such as in the Digital Humanities, span a wide range of activities and, notably, also languages, which rely in part on advances made in tool development in the Natural Language Processing (NLP) field. An example is the digitisation project of the notebooks in the |xam and !kun languages of Southern Africa as part of the Llarec project (Digital Bleek and Lloyd). Similar projects take place on other continents, such as the development of digital tools for the Dongba pictogram language (southern China) (Xu), and Ottoman Turkish (Kirmizialtin and Wrisley). Neither language is in the same league as English when it comes to OCR tools for digitisation, for instance. In fact, tooling for the languages was minimal to non-existent, hampering progress in research and software implementation projects because the infrastructure hurdles must be overcome as well before carrying out the intended task.

Many terms are used to describe human languages according to the data resources or tools for the development of their human language technologies and the applications that require them. In the context of African languages, terms like under-resourced, resource-scarce, less-resourced, low-resourced, or low resource languages (LRL)[1] have been used to characterise the lack of data resources, which preclude or slow down the development of computer-based and data-driven technologies in these languages and makes data-intensive NLP an inappropriate approach. Furthermore, there is an assumption that all African languages are in the same category of LRL, while conversely a similar assumption obtains, that English and other European languages are high resourced languages (HRL) and that LRL is necessarily only relative to HRL. In this paper we provide a deeper understanding of this complex phenomenon.

The reference to African languages as LRL (Bosch et al.; Pretorius & Bosch; Keet & Khumalo) does not necessarily apply the same way to all the African languages. To appreciate this complexity, we need to understand what language resources entail. From the computational side, the NLP literature refers to LRL as having limited online corpora and computational grammars. Alternatively, it is formulated in terms of missing something, e.g., "[…] lacking large monolingual or parallel corpora and/or manually crafted linguistic resources sufficient for building statistical NLP applications"[2], as if statistical NLP is the benchmark, or "LRLs can be understood as less studied, resource scarce, less computerized, less privileged, less commonly taught, or low density, among other denominations" (Magueresse et al. 1). Compared to what? Less than which other language? For instance, Dutch is a well-resourced, but 'less' computerised, language and has much fewer speakers compared to English. Clearly, following these measures, then every language other than English falls in the 'less than' or 'low' resource category. These indications are coarse-grained and do not assist in categorising which language is where on the spectrum. To that end, several attempts have been made by counting a selection of resources and devising a metric for classification, such as counting labelled and unlabelled corpora (Joshi et al.) and availability of data and tools for a list of NLP tasks (Berment; Krauwer). But is the combination of the number of resources and their availability the only dimension of a low-resourced language? We argue that it is not.

---

[1] Henceforth, we use LRL to mean Low-Resourced Language; see the appendix for the deliberation of the terminology.

[2] From a Medium article by (Naumann), which was likely lifted from Tsvetkov's lecture slides (Tsvetkov 26).

This is illustrated in various half-measured attempts to bridge the gap between the LRL and English. In international collaborations, for instance, the attitude is often one of "just get the translators" to create the parallel corpus between language x and English or "quickly list how adjectives work in your language" for a grammar-based approach to spellchecking or natural language generation. This stance obtains in print as well (e.g., (Hämäläinen)). What transpires in practice is that, even with money available, there may be insufficient translators available, or they are not specialised in the domain one needs the translations, or they are not translators by training. Or when there are enough people available to scan text, the OCR software works 'best' only for the language it is optimised for[3], which is not the language being scanned and so the scanned text is riddled with errors. One then either has to start over manually or find a computer scientist to develop language-optimised OCR[4]. Both are setbacks and cause further delays in trying to close the gap. Developing the materials needed for a grammar-based approach may require a full literature search and compiling content from different, often outdated, sources that may not have been tested on a sizeable number of words, rather than having the luxury of reaching for a grammar book that only needs to be fetched from the bookshelf and digitised. Based on the foregoing, there are more facets to LRLs than simply not having a lot of digital language material.[5] The so-called

[3] Good OCR software does not only try to recognise the characters, but also add probabilities to words and characters that it is not sure of. This may be 'fixing' a string so that it passes a spelling check for the language it is optimised for (e.g., changing the isiZulu *wena* into the English 'wend') up to analysis of the sentence to choose among alternative corrections (e.g., if the sentence misses a verb, then the candidate fixes for the unclear string will select a verb).

[4] This meanwhile has been done for the 11 South African languages (Hocking & Puttkammer) – after the IsiZulu National Corpus development was delayed due to such scanning errors (Khumalo).

[5] The three examples were not merely theoretical, but they have occurred to the authors.

'lived reality' of working with LRLs is more complex than limited data or simply missing digitised corpora and grammars.

The imperative exists, therefore, that we should have a comprehensive understanding of the broader context. This enables researchers to appreciate what the presence or absence of a particular resource entails in the characterisation of a particular language, for LRL and HRL alike. Acknowledged, the characterisation of HRLs utilises a similar sweeping generalisation trend. The assumption that all European languages are as highly resourced as the English language is flawed, especially in the absence of a clear matrix that assists in such a characterisation.

In this paper we focus on three key issues:

1. What are really the distinguishing characteristics of LRLs (and, by extension, 'non-LRLs')?
2. What are the characteristics of levels of resourcedness?
3. Which language fits where and why?

To answer these questions, we first devised a list of 11 key dimensions for languages and then assessed which of those applies to which level of resourcedness. This characterisation provides a more comprehensive understanding of the multi-faceted concept of resourcedness of a language beyond the counting of speakers, tools, and corpora, with a focus on teasing out details of 'low' resourcedness. We then took the classification and allocated all 11 official South African languages, as well as several other African languages and languages spoken elsewhere in the world, demonstrating that the classification is operationalisable. It may further assist, e.g., language policy planning.

Section 2 discusses typical descriptions of LRLs used in the literature. Section 3 introduces the matrix used for characterizing language resources and analyses the features assigned to the various categories in the characterisation of languages as Very LRL, LRL, RL and HRL. It will be shown that while most (European) languages are classified as HRL, English is leading as a Very HRL, a characterisation that hitherto did not exist. We discuss in Section 4 and conclude in Section 5.

## 2. Related work

In the fields of NLP, computational linguistics, and human language technologies, i.e., from the computing side, the aforementioned 'lacking …' view on LRLs is often repeated in various wordings across publications, where research papers adopt such descriptions from survey papers and possibly from online media as well[6]. Besacier et al.'s survey describes LRLs as "a language with some of (if not all) the following aspects: lack of a unique writing system or stable orthography, limited presence on the web, lack of linguistic expertise, lack of electronic resources for speech and language processing, such as monolingual corpora, bilingual electronic dictionaries, transcribed speech data, pronunciation dictionaries, vocabulary lists, etc." (Besacier et al. 60), which is roughly repeated in Ranathunga et al. and from both surveys to hundreds of research papers. Besacier et al. based their description on the Basic Language Resource Kit (BLARK) "as the minimal set of language resources that is necessary to do any precompetitive research and education at all", described in (Krauwer 11), and a similar approach based on counting resources by Berment. Berment distinguishes between low, medium, and well-resourced, and provides a list of types of resources – dictionaries and a range of software applications for multiple NLP task – to assist counting them for a given language, which is then used to compute a value for categorisation (Berment 20-22). Krauwer provides a longer list of language- and speech technology tasks and modules with data and application availability and their importance for the NLP task, such as parsing and referent resolution (written language) and prosody and pronunciation lexicon (speech) on corpora and information access applications. The availability status of the resources collected is then estimated on a 10-point scale and added up so that eventually the resourcedness of languages may be compared.

Joshi et al. took a narrower scope on resourcedness assessment than Berment and Krauwer, by counting the resources available on the ELRA map for labelled dataset, the LDC catalog for corpora, and Wikipedia pages as an indication of unlabelled data in order to devise six levels of resourcedness which is, to the best of our knowledge, the most fine-grained list on resourcedness of languages. They gave them the descriptive names of the left-behinds (n=2191; a.o., Dahalo), scraping-bys (n=222; a.o., Cherokee, Fijian), hopefuls (n=19; a.o., isiZulu, Irish), rising stars (n=28; a.o., Indonesian, Afrikaans), the underdog (n=18; a.o., Russian, Dutch), and the winners (n=7; a.o., English, Japanese). Given an estimated 7000 spoken languages, there is an implicit 7th category behind the left-behinds for the remaining 5000: the 'not even considered'. Assessing their taxonomy[7], the latter group includes, among many, Mpiemo (for which a POS tagger does exist (Hammerstrom et al.)), Pokomo, and Sanga of the Niger-Congo B ('Bantu') family of languages. This also illustrates a shortcoming of the 'counting resources' approach to defining and categorizing LRLs: there are always resources that are overlooked, which is due to a range of reasons interfering with the resource collection efforts and how searches for resources are conducted. For instance, multiple languages have several names or spelling variants, and their use may be buried in a data collection paragraph in the paper and thus missed when only titles and abstracts are scanned in the harvesting stage (Keet). Detecting mentions, like Joshi et al. did, will have missed a paper on Mboshi/Embosi, for the language is listed in their taxonomy only as Mbosi without spelling variants.[8]

[6] The Medium article (Naumann) and (Tsvetkov 26) are roughly repeated in the surveys cited. In contrast, (Bender) only characterizes high resource languages, leaving the 'low' implicit as not having all that.
[7] https://microsoft.github.io/linguisticdiversity/assets/lang2tax.txt

The approach of counting a selection of resources of LRLs, and, hence, the implicit lack of things, is a recurring theme for categorising LRLs, although at times LRLs are informally described slightly more broadly. Hedderich et al. broadens LRL as an "umbrella term" that "covers a spectrum of scenarios with varying resource conditions", which include "threatened languages", languages that are "widely spoken but seldom addressed by NLP research", and "non-standard domains and [NLP] tasks" (Hedderich et al. 2545). Yet, then they narrow it down when teasing out aspects of resourcedness with their three dimensions: of availability of labelled data relevant for the task at hand, of unlabelled language or domain-specific text, and of "auxiliary data" such as a relevant knowledge base in the target language to enhance the NLP or a machine translation system to generate training data. A key observation is that indeed there are dependencies across NLP tasks and that absence of such auxiliaries hampers progress of the intended task. Occasionally there are hints at more features of LRLs beyond data and software tools, as in "less studied, resource scarce, less computerised, less privileged, less commonly taught, or low density, among other denominations" (Magueresse et al. 1, based on Cieri et al.), but it is then still narrowed down for their works to refer to "languages for which statistical methods cannot be directly applied because of data scarcity." (Magueresse et al. 1) or "those that have fewer technologies and especially data sets relative to some measure of their international importance." (Cieri et al. 4545).

Within the South African context, the characterisation that emerged from the two Human Language Technology (HLT) Audits in 2011 and 2018 (Moors et al.; Sharma Grover et al.) is the one that refers to African languages as generally "under-resourced" or "lesser resourced". This is understood to mean that these languages have insufficient resources, as discussed in Section 2. The resource types covered in these audits range from the text/speech corpora to speech-to-speech translation systems.

[8] The list of detected papers has not been made available by Joshi et al., but by their methodology, it would have missed at least the LREC2018 paper on Mboshi, being Rialland A, Adda-Decker M, Kouarata G-N, Adda G, Besacier L, et al. Parallel Corpora in Mboshi (Bantu C25, Congo-Brazzaville). 11th Language Resources and Evaluation Conference (LREC 2018), ELRA, May 2018, Miyazaki, Japan.

Concerning categorising languages other than explicitly by the types of resources for computational use, there are ample categorisations. The one that comes closest to a combination is Kornai's "digital language death" assessment (Kornai 1), where the author makes an argument for a digital divide, itself a broad dichotomy of well-resourced languages accessed by every user of a computer device, and the other end of the spectrum with languages that have no Wikipedia, no data, no forms of technology support. In his thesis, languages are digitally ascending or not, which was based on his machine classified languages, having devised a sampling of prototypical members of each category to seed the clusters, and a 5-point scale – digitally Thriving, Vital, Borderline, Heritage, and Still (Kornai 5-9). The scale was inspired by the 12-level EGIDS classification to assess language endangerment (Lewis and Simons 110). Other such language vitality and endangerment levels have been proposed and is extensively discussed in (Lewis and Simons). How exactly language endangerment and resourcedness interrelate is yet to be investigated in more detail. Anecdotal evidence suggests that they do not go in lockstep; see also, e.g., (Hämäläinen).

In sum, the literature provides neither a clear definition of what LRLs really are, nor what exactly their characteristics are beyond the 'lack of' and 'less' resources for computation. The following section provides a detailed account of the conceptual range of resourcedness of natural languages.

## 3. A more comprehensive view on low-resourced languages

While there are 'few' resources for LRLs, those few resources are not all alike and the LRLs are embedded in a broader context that makes the efforts of getting out of the 'low-resourced' status also not all alike. For instance, aspects of language vitality play a part when recruiting workers for annotating text or for a human evaluation of algorithmically generated text. Further, it requires a characterisation of its own as compared to reducing it to the negation of high-resourced, especially since, by number of languages, being low-resourced is the norm and high-resourced the exception. To this end, we first compiled a list of dimensions of what low-resourced and what high- or well-resourced entails, which is described and illustrated in Section 3.1. Second, we allocate applicable dimensions to each of the five categories from very low to very highly resourced and to what extent they apply and operationalise that for a section of languages, which is described in Section 3.2.

### 3.1 Dimensions of resourcedness

The list of dimensions is summarised in Table 1 and elaborated on in the remainder of this section in that order. The first eight dimensions are described in no order of priority and focus on multiple key aspects and consequences related to LRLs. They are a constellation of varied dimensions, not solely of the corpus-and/or-tool-counting, nor solely on societal aspects of language vitality, for the two are interdependent when one actually works with low- and high-resourced languages.

Elucidating these dimensions led to the question: 'what do high, and very highly resourced languages have that one would like to have for LRLs too?'. The aim here is to commence bringing those benefits or features to the fore, which are summarised in features 9-11. We elaborate on each one in turn.

*Limited grammar* refers to the situation where a limited amount of the grammar is known, the limited documentation available may be outdated, and the few rules known may not have been evaluated on whether they hold generally or always or have exceptions, and this exists mostly or only on paper, rather than as a digital resource. It also includes the case where there is no grammar reference book for the language in question, be it an outdated book or a recent book. What this practically means is that any grammar feature that one wishes to use computationally requires careful investigation through consulting various scientific literature, textbooks, or colleagues to distil what the rule(s) may be, and a certain amount of testing with sample data to determine whether the candidate rule(s) hold. If so, then it is still to be computerised, which may add a new hurdle of compatibility challenges with the implemented grammar framework. For instance, the popular academic framework of Universal Dependencies (de Marneffe et al.) may not be readily usable for the LRL in question because it relies on the lexicalist paradigm whereas the language may need annotations at the sub-word level. If so, then one first needs to investigate an extension, implement software support for that, and only then get to the actual task of computerising the grammar; an example of this detour is presented by Park et al. for St. Lawrence Yupik. It may be possible to devise a scale for this dimension for a sub-classification, involving the properties of no versus outdated documentation, no versus sparse versus occasional scientific publications in linguistics venues, and an attribute on 'special' language features that do not occur, or are not considered, in the medium and high-resourced languages.

*No/very limited corpora (<100K tokens)* concerns both monolingual and parallel corpora with another language. The most challenging position to start from is when there is very little written text at all. Since the emergence of social media,  this will be less likely the case, but such data requires more effort to create a corpus due to code switching in text messages and privacy concerns.  While creating a corpus is a relatively straightforward task, resources are needed to carry it out and sustain it: people to do it, funding to pay them, a plan for longevity, and open access to the corpus. Common first steps to putting a corpus together are the freely available bible translations, no matter the datedness—as if the LRL would not have developed over 100 years—and in a very specific genre that is typically different from the target genre, since when there are no to very limited corpora, genre matters less. An experiment by Ndaba et al. showed that even training on a tiny corpus of about 20000 tokens from current news items already outperformed a language model trained on the bible in developing spellchecker for modern isiZulu. Then there are the OCR challenges in digitisation, as mentioned in the introduction (footnote 3).

*Little basic HLTs*. With 'basic' HLTs we mean entry-level digital support features that facilitate the writing of text by its speakers, such as spellcheckers, an autocomplete function for tools like WhatsApp, and an online bi-directional dictionary between words in the LRL and those of a more widely-spoken one. Conversely, a misaligned 'autocomplete' that adds errors in the text message obstructs the use of the LRL and slows down writing, as do characters not in the 26-letter alphabet of English if there is no interface adapted to the LRL's alphabet (Lackaff and Moner), therewith discouraging users to write in that language. Technologically, techniques to provide such functionality may be highly sophisticated, yet for the HLTs they are perceived as part of the core enabling infrastructure.

The 'little basic HLTs' may thus be refined further in a classification with parameters in the order of: language-relevant keyboard, language-specific autocomplete, recognition of named entities and common names for people and places, autocorrect spelling checker, online monolingual dictionary, online bi-directional dictionary to a relevant other language, and rudimentary grammar checking (e.g., noun class agreement, verb inflection).

*Few people drive the advancement of the language*. This is sub-divided into few researchers in languages and linguistics in the LRL as well as few researchers in computer science and engineering that have the language as a research focus in their HLT, NLP, or computational linguistics specialisation, few companies that make it their business, and insufficient language teachers at especially primary and secondary schools. The shortage of researchers hinders progress not only in language understanding but also in prospects to devise novel algorithms to develop HLTs that work for the language's peculiarities, since the international tools typically either have to be adapted or started from scratch (e.g.,

Steinberger et al. and De Pauw and Wachaga, among many). The presence of few HLT people in industry indicates that the language may not be economically profitable even though many people may speak it. The shortage of language teachers in schools typically entails that learners in higher grades may not have their home language as language of teaching and learning, and consequently they will not learn their mother tongue (home language) well. This, in turn, has a knock-on effect on students entering tertiary education, for being relatively behind in language knowledge, which means either the degree programme will contain less advanced content compared to, say, the English programme at a university in the UK, USA, Australia, or New Zealand, and thus results in either a comparatively less-skilled language graduate or the programme is comparatively harder due to the need to catch up, which leaves fewer students to choose such a degree and more to drop out. This reduces the pool of candidates for academic posts to conduct research into the language or to teach a new generation of language teachers. It is a vicious circle at best and a downward spiral at worst. Additional effects of limited language education are that it makes it harder to recruit competent participants in evaluations for one's research and sub-competent evaluators add noise to the development of algorithms to develop the tools that may then need another round of editing and evaluation, and thus overall taking up more effort and time to develop than required for RLs and HLRs.

*Little/no funding* concerns all the dimensions. It entails that most advances are made by passionate volunteers, who have different incentive mechanisms and may have different preferences for what HLT to invest time in than the few, if any, funders may have. It also affects the longevity of the outputs produced in a project: when the limited funding comes to an end or the volunteer's initial enthusiasm runs out, resource maintenance suffers (e.g., tool compatibility, the domain name not renewed) and will slide into disuse, for someone else to re-do the same or a similar effort later for lack of availability of the preceding outputs. This keeps a language going in circles with basic HLTs, rather than advancing the state of the art. An example are the isiZulu spellcheckers: they were developed in the 2000s (De Schrijver and Prinsloo) but the tool and code were never released, developed from scratch again as a plugin to OpenOffice (Dwayne) for older versions (before 4.x) and therewith became defunct in the early 2010s, created as a stand-alone jar file in 2016 based on new algorithms of (Ndaba et al.), and implemented again in 2022 by CTexT as a plugin to the Microsoft Office Suite on the PC only (SADiLaR). Categorising on 'limited funding' therefore concerns whether there are earmarked funds at all, and if so, the amount relative to the costs where the activities take place and whether that is for research, innovation and technology development, or scraping by on essential tasks of teachers and teacher training.

*Limited education*. This relates to the previous discussion on few teachers in praxis, but here we refer to the education system itself and the programmes that are available. Specifically, on whether the language is used as language of learning and teaching, or whether one can take it as a subject at school as an 'additional language' subject, and whether it is even possible to study it as a degree at a university. For instance, the University of Cape Town, South Africa, only offers isiXhosa as a major, but there is no full BA in languages, literature or linguistics taught in isiXhosa (UCT), and the information is provided only in English, which is a similar situation at other South African universities for other languages[9]. Compare this to, e.g., studying the minority language Irish at Trinity College Dublin, Ireland, where the information page is already in Irish and there are two 4-year degree programmes (TCD1; TCD2). A lower bar to pass is attitude to languages, where there may be "language centres" at an educational institution to learn multiple languages at conversational and certificate levels as extracurricular activities; e.g., the small Italian university Free University of Bozen-Bolzano's language centre offers courses in nine languages[10] and Wageningen University, the Netherlands, offers language courses in 15 languages[11]. In contrast, the University of Cape Town's language centre only offers several versions of English[12]; the University of Stellenbosch offers English, Afrikaans, and isiXhosa[13]. For LRLs, even such options thus may be limited to non-existent.

*Limited/no government policy framework*. While there is a difference between having a language policy framework and actually implementing it, no policy at all is worse. In the South African context, the Constitution provides for the development of the 11 official languages in order to achieve "parity of esteem" between them and various language policies have been promulgated to achieve this. However, these have been criticised as "[…] pious articles of faith, without any force nor any effect" (Khumalo as quoted in Anstey). It remains to be seen how the Department of Higher Education and Training (DHET) will provide for the implementation of its Revised Language Policy Framework for Public Higher Education

[9] For instance, the isiZulu majors at the University of KwaZulu-Natal https://soa.ukzn.ac.za/clustersdisciplines/language-and-literature/isizulu/

[10] https://www.unibz.it/en/services/language-centre/ it offers Arabic, Chinese, English, French, German, Italian, Ladin, Russian, and Spanish.

[11] https://www.wur.nl/en/education-programmes/wageningen-into-languages/language-courses.htm It offers Arabic, Chinese, Dutch, English, French, German, Italian, Japanese, Korean, Norwegian, Portuguese, Russian, Swedish, Spanish, and Turkish.

[12] https://uctlanguagecentre.com/

[13] https://languagecentre.sun.ac.za/

Institutions, which took effect in January 2022. Thus, this dimension may be interpreted as a scale, including: no policy framework, a (current or outdated) policy, implementation plan, implementation plan (partially or fully) being implemented, and the various levels of success or failure of implementation.

*May not have a monolingual dictionary*. We distinguish between bi-lingual dictionaries that provide English/French/Spanish/Portuguese/etc. translations to the LRL vocabulary and the recursiveness of a monolingual dictionary. The latter is seen as a measure of a certain level of resource development, status, and use, in that the vocabulary indeed can be explained in the language itself rather than having to resort to approximations with a foreign word. In other words, the LRL is thus expressive enough to describe one's language in one's own terms. This, perhaps, also concerns a legacy of colonialism and acting against it: indigenous languages were assumed to be not full-fledged languages because a people speaking it lived 'primitively' in the eyes of the coloniser and a monolingual dictionary then becomes a statement of belonging to humanity on par with other peoples. It also requires more effort to develop a monolingual dictionary compared to word lists, and it entails that there are at least some people driving the language forward and that there is a notion of standardisation of the language and therewith institutionalisation. Note that not having a monolingual dictionary does not imply that the language is not expressive enough. While this dimension can be treated as a Boolean yes/no variable, gradations may be identified, such as no dictionary, one under development, early partial attempts shared with the community, a reasonably complete dictionary, and a widely circulating complete and current monolingual dictionary that is being maintained.

*Choice among computational grammars* is the first dimension/feature worded from the perspective of better resourced languages. This dimension of being able to choose a computational grammar for one's need or preference includes both choice between grammar paradigms (e.g., dependency versus phrase-structure grammars) and within-paradigm amongst specific frameworks, like UD (de Marneffe et al.) versus SUD (Gerdes et al.) dependency grammars. It also entails that these computational grammars are sufficiently comprehensive and perform well on the relevant quality metrics.

*Large corpora, with genres*. From an LRL perspective, a corpus of 1 million tokens of any genre of corpora that can be found is impressive; for better researched languages, it would be in the order of at least 500 million to 1 billion tokens. In addition, such corpora would be separated by genre, such as news articles or novels, and by language varieties by geographic region or time period, such as American English versus British English or modern French versus renaissance French.

*Many tools and choice among tools, on multiple devices and OSs*. Unlike the intermittent serial spellcheckers for isiZulu illustrated above, there is an abundance of choice of tools for the same operating system as well as across software, such as a spellchecker not only for MS Word or only copy-paste into a Web app and copying the corrections back in, but it is embedded in the tools upon installation, regardless of the device, the operating systems, and the writing software. For instance, the tool or plugin works on both Android and iOS smartphones, in browsers when writing an email, a blog post or a lesson page in a learning management system, in the Thunderbird Desktop email application on Windows, MacOSX, and Linux or even all platforms, and any popular text editor. In other words: there are multiple tools that offer the same functionality across that spectrum. This also includes a choice of tools for NLP tasks, such as the Stanford Parts of Speech (POS) tagger and the POS tagging module that comes with Natural Language Toolkit (NLTK), among many. This may be sub-categorised, or a scale defined, regarding only resources (tools) for core or enabling functions versus also advanced tasks, and whether the tools are at the proof-of-concept, prototype, or end-user usability level of maturity.

*Table 1. Summary of the dimensions and components thereof, where applicable.*

| Number | Topic | Component-topic |
|---|---|---|
| 1 | Limited (written) grammar | |
| 2 | No/very limited corpora | |
| 3 | Little basic HLTs | |
| 4 | Few people driving the advancement of a language | 4a. researchers<br>4b. industry<br>4c. teachers |
| 5 | Little/no funding | |
| 6 | Limited education | 6a. as LoLT<br>6b. as additional language at school<br>6c. as degree offerings at university |
| 7 | Limited/no government policy framework | |
| 8 | May not have a monolingual dictionary | |
| 9 | Choice among computational grammars | |
| 10 | Large corpora, with genres | |
| 11 | Many tools and choice among tools, on multiple devices and OSs | |

This concludes the presentation of the dimensions. It is not inconceivable further dimensions may be added, if broader operationalisation requires it. The criteria or dimensions characterised above form a matrix that we will use in Section 3.2 when analysing the complexity of understanding LRs, where we will show that the characterisation of languages from Very LRL to Very HRLs is based on well worked-out criteria or dimensions, with examples of such languages provided.

### 3.2 Analysis

The analysis uses feature-based criteria. The features are assigned to each of the five categories Very LRL, LRL, RL, High RL and Very HRL in Section 3.2.1. The 11 official languages of South Africa are assigned to categories in Section 3.2.2 together with other African and global languages and illustrated in more detail for isiNdebele.

#### *3.2.1 Matching dimensions to the level of resourcedness of a language*

A basic categorisation, or levels, of resourcedness easily adjusts toward a familiar 5-point Likert scale version. For resourcedness of languages in the broader context, that is, first, the distinction between well-known "very LRL" and "LRL". What comes after is also not one homogeneous group, and with an averagely resourced language ("RL") in the middle, the matching other side it is high-resourced language ("HRL") and the extreme on that end, a very high-resourced language ("very HRL"). We assign the features to the five levels as follows:

- Very LRL: 1, 2, 3, 4, 5, 6b, 6c, 7, 8
- LRL: 1, 2, basic 3, 4a, 4b, sparse 4c, 5, 6b, 6c, 8
- RL: 3, 5, may have one of {9, 10, 11}
- HRL: some 9, some 10, some 11
- Very HRL: 9, 10, 11

This is also visualised in Table 2. What is especially clear from the visualisation, is that the difference between LRL and RL seems to be the largest. 'Crawling out of the pit' of relative disadvantage to 'getting by fine' may indeed well be the hardest to do, since once the system is rolling, it makes it easier to build upon it and move forward faster but getting it rolling takes more effort due to the many dependencies of functionality.

It also shows that "Limited LOLT" (6a) turned out to be more nuanced than a blanket yes or no for a whole category. The (very) LRLs are used especially in primary schools, be it officially or in practice and this happens as well in the resourced languages. It exists as a dimension, because the LRL is often perceived as lower status and replaced by a colonial language as LOLT especially in higher grades (Grade 4 onwards) and better schools, but this does not hold for all LRLs. Hence, it can be applied only to a particular LRL or very LRL, rather than the category.

*Table 2. Visualisation of the dimensions by level of resourcedness. X: yes/applies; v: applies to a considerable extent i: applies to some extent; blank cell: does not apply.*

| | 1 | 2 | 3 | 4 | | | 5 | 6 | | | 7 | 8 | 9 | 10 | 11 |
|---|---|---|---|---|---|---|---|---|---|---|---|---|---|---|---|
| | | | | a | b | c | | a | b | c | | | | | |
| Very LRL | x | x | x | x | x | x | x | | x | x | x | x | | | |
| LRL | x | x | v | x | x | v | x | | x | x | | x | | | |
| RL | | | x | | | | x | | | | | | i | i | i |
| HRL | | | | | | | | | | | | | v | v | v |
| Very HRL | | | | | | | | | | | | | x | x | x |

*3.2.2. Categorising languages*

With the aim towards operationalising the dimensions and levels of resourcedness and illustrating the categorisations and 'lived reality' better, we now turn to examining which language can be categorised where and why. We do this specifically for the 11 official languages of South Africa, several other more or less well-known languages in Africa, and a selection of other languages in the world for comparison and to provide ample examples in each category. There are many ways how such a categorisation can be done, which each have their own shortcomings. For instance, one may try to collect all the resources and tools ever developed for the language or only those that are functional (be it assessed or claimed on a paper or open access) or collect papers about resources from a wide range of publication venues. Such resource collections and audits have been carried out, including, the HLT audits of South African languages resources (Moors et al.; Sharma Grover et al.), the South African Digital Languages Resources repository (SADiLaR repo), Masakhane (Masakhane) and Lanafrica (Lanafrica) repositories, and the articles in top conferences only (Joshi et al.), as discussed in Section 2. Many advances for African languages do not reach the top conferences, however, for various reasons. The HLT audits and the SADiLaR repository are limited to South African languages only and only those tools developed with SADiLaR funding, Masakhane is limited to corpora only, and Lanafrica is collecting datasets only since 2022 and does not include tools. That is: relying on only any one source or strategy for data collection to categorise languages on their tooling resourcedness is substantially incomplete and therewith will present a distorted view. Even if they were to be complete or an ensemble collection method were to be used, it then would need to be converted into a scale to measure, and it still would not indicate anything about aspects such as tool maintenance, quality, and the broader context and the features of written grammar, funding, people, or education. Conversely, an expert-based classification using the aforementioned features remains opaque at least in part due to incomplete evidence in that case. Data for the features may be found to substantiate a categorisation value, but some of them have inherently fuzzy boundaries. For instance, evidence of teacher shortage can be obtained from a Department of Education, but a teacher shortage of 20% is different from a shortage where only 1 in 10 schools offer classes in the language to be categorised, but so are 25% and 1 in 9, respectively. More research is needed to refine some of the values for the features to enable a more robust categorisation.

Meanwhile, as a preliminary operationalisation assessment, we use the expert-based approach, availing of the fact that the authors are based in South Africa, are knowledgeable about several African languages elsewhere on the continent through regional research, service, and outreach, and knowledgeable about several other languages overseas thanks to background, education, research, and being active members of the international academic community.

Let us take isiNdebele, a language spoken both in South Africa and Zimbabwe, which exhibit some variations among the two geographic locations due to the border limiting interaction and influences of co-located and adjacent languages, these being mainly Venda in South Africa and ciShona in Zimbabwe. It has an estimated 2.6 million first-language speakers in Zimbabwe and another 1.1 million in South Africa[14]. There are newspapers, TV news bulletins and so on written and presented in isiNdebele.

Grammars for the two varieties are limited. This is because for a long time they both relied on the *Textbook of Zulu Grammar* (Doke) and *Uhlelo lwesiZulu* (Nyembezi), which is very limited in terms of its linguistic description. Recent grammar descriptions are still based mainly on the two isiZulu grammar books.

Corpora: A first corpus of isiNdebele spoken in Zimbabwe was developed under the auspices of the ALLEX Project (ALLEX) in 2002 but is not online accessible (Hadebe), a Southern Ndebele corpus of 2730 words is available from (Wortschatz Leipzig) and SADiLaR lists various corpora for TTS and multilingual wordlists such as for COVID-19 and grade R-6 mathematics terms (SADiLaR Search), and the annotated NCHLT corpus of 41013 words (NCHLT isiNdebele).

[14] The various censuses consulted are around 10 to 20 years old, which are also estimates. The actual number is likely to be higher.

Basic HLTs: An isiNdebele spelling checker is claimed to exist for MS Word in 2022 (see footnote 10) and an older, inaccessible one is listed in the SADiLaR repository (isiNdebele spelling). There are currently no autocomplete, grammar checkers, morphological analysers, or POS taggers, hence, there are also no tools that use these functions.

Few people: Because of the limited availability of programmes that teach isiNdebele at schools, colleges, and universities, there are less than ten thousand (<10 000) researchers that have been trained in these languages on various aspects of linguistics. Even fewer (<100) are active researchers.

Additionally, funding remains a huge challenge as these language varieties have been largely unable to attract funding for digital development from government or industry. The only notable international collaboration to fund the development of corpora and dictionaries was through the ALLEX project. In South Africa, isiNdebele receives very limited funding from the Pan South African Language Board (PanSALB) for terminology development and dictionary making.

Limited education: The University of Mpumalanga offers a major and an BA Honours in isiNdebele, as does the Great Zimbabwe University, and a few other universities offer individual modules. A huge setback is the recent scrapping of language grammar modules for isiNdebele and ciShona at the University of Zimbabwe, a development that sees the reversal of the advances made through the ALLEX Project in the late 1990s. Following the introduction of the Education 5.0 framework for higher education institutions in Zimbabwe, the University of Zimbabwe developed its strategic plan (2019 – 2025) premised on the foundational principles of Education 5.0. According to (Muyambo and Ndamba 7), because of this "[…] no module was being taught and learnt in indigenous languages despite their agreement that the mother-tongue is important in the teaching and learning processes of learners".

Limited/no policy framework: For South Africa, a policy exists insofar as it exists for all nine African languages, as discussed above (it exists on paper but is sparsely implemented if at all). The Constitution in both Zimbabwe and South Africa demonstrate a constitutional imperative to elevate these languages for use in all spheres of life. What is lacking is the implementation strategy and framework (backed by direct funding) to execute the spirit and letter of the constitution.

May not have a monolingual dictionary: Both varieties of isiNdebele have monolingual dictionaries. However, specialised dictionaries are sparse, with only one music dictionary (Nkomo and Moyo) for isiNdebele spoken in Zimbabwe. Efforts to develop similar specialised dictionaries in other disciplines is hampered by both funding and very limited human capacity both in terms of expertise and the number of people available.

In sum, IsiNdebele in both Zimbabwe and South Africa remains a Very LRL. This is because the available grammars are dated and limited (dimension 1); the corpora are very small (less than a million for the ALLEX Ndebele corpus) (dim. 2); there are no HLTs for Zimbabwean IsiNdebele and an inaccessible spellchecker for IsiNdebele in South Africa as mentioned above (dim. 3); there are <100 known active researchers in the two language varieties (dim. 4); funding is often linked to other (relatively bigger) African languages (Shona in Zimbabwe in the case of the ALLEX Project) and other 9 official African languages in South Africa in the case of the Department of Science Technology & Innovation (DSTI) funded SADiLaR (dim. 5); relative to dimensions 6 a, b, c, there has been a reversal of the use of IsiNdebele as a language of teaching and learning, and as taught subjects following the introduction of Education 5.0 framework and there is evidence of limited offerings at Honours and post graduate levels; and dimension 7 applies as well. While dimension 8, on the monolingual dictionary, is partially met, the others are all on the lower end of their respective scales, and therewith thus remaining in the Very LRL category.

For the Very LRL and official languages in South Africa, we classify Xitsonga, Venda, Sepedi, siSwati, isiNdebele, and Sotho as such. Other examples from Africa, among many, are Mboshi and Malagasy. The LRL category may be most interesting theoretically. IsiXhosa from South Africa clearly is solidly LRL. Concerning tooling, but even more so political clout and *lingua franca* practices, one might wish to argue for a sort of 'upper LRL' that aligns with the "hopefuls scraping by" in Joshi et al.'s terminology, being isiZulu and Setswana in South Africa and Kiswahili in Tanzania and Kenya. They are still LRL but have a few more resources than isiXhosa. As RL, there are Afrikaans and South African English for South Africa, as well as Portuguese and Arabic in Africa, and, e.g., Italian internationally. Examples of HRL are Dutch, German, French, and Mandarin. The only language that is very HRL is English.

## 4. Discussion

The use of the term low-resourced in the literature has hitherto been problematic. The views adduced from Hämäläinen – that in the NLP literature Chinese (1.2 billion speakers), Arabic (422 million speakers), Bengali (228 million speakers), Japanese (126 million speakers), Vietnamese (76 million speakers), Dutch (24 million speakers), Sinhala (17 million speakers) and Finnish (5 million speakers), have had several tasks referred to as low-resourced – is, at best, based on a subjective personal scale or with English being the benchmark. For the term low-resourced to carry any scientific currency in appreciation of research and application projects that avail of NLP tasks more generally, and indeed in other specialised environments, the exigency exists to define in a scalable way the range of language resources.

We have argued for 11 dimensions for resourcedness to address the semantic obfuscation presented by the current use of the term 'low-resourced'. The 11 dimensions were matched to five groupings of resource characterisation towards a familiar 5-point Likert scale: VLRL; LRL, RL, HRL, VHRL, as presented in Section 3.2.1. These categories provide an answer to the problem raised by Hämäläinen, that of lumping together all the languages other than English as LRL, which he rightly observed as incorrect. LRLs are not a homogenous group, and the 5-level categorisation provided here means that more languages can be allocated into these categories in a more meaningful way. This may also assist with "LRL sessions" at NLP conferences, to clarify what experiences and research are asked for, and for authors to position their contributions more precisely. Likewise, not all projects involving natural language resources will have the same pace of progress due to the different amounts of NLP tasks entailed in the overall project, and the prerequisites they assume that have to be met before one can go start investigating research question from, say, sociology or history, or create software applications in the target language as part of, e.g., an ICT4D project such as software-managed agriculture.

The categories as defined from the 11 dimensions are expected to be useful in providing a language policy and planning framework for the people responsible for the development of language technologies for languages that are in the levels VLRL and LRL. The deployment of resources and instruments required to develop them can henceforth be based on a scientific criterion, instead of either population demographics alone or hastily discovered or cherry-picked tool, corpus, or research paper-based counting. It has been shown that the fact that a language has fewer speakers (e.g., Finnish) or is an endangered language (e.g., Skolt Sami) (Hämäläinen) does not mean that it is a LRL. The categorisation will therefore provide government and language policy experts a further mechanism in developing more informed language policies and more effective language implementation frameworks that address the characteristic elements of language category as articulated in the eleven dimensions. Such policies and frameworks may need to set feature cut-off points appropriate for their context, such as about the comparative number of language teachers in schools. This also may lead to insights into why one language remains (very) LRL whereas another may reach RL, HRL, or Very HRL. For instance, the extent of the effects of speakers organising themselves, such as *Afrikaanse Taalraad* or *Academie Francaise*, on the amount and quality of resources, or a retrospective assessment of relevant university degree offerings and their enrolments on the level of resourcedness of a language. The dimensions proposed here can assist focusing the research on what enables or impedes the development of language resources.

The categories thus also provide a rubric for planners to track the progress in the development of each language through the categories VLRL; LRL; RL; HRL and VHRL. This is particularly useful within the South African context, as the South Africa Constitution clearly articulates the commitment to develop all the 11 official languages in order to achieve "parity of esteem" between all the eleven official languages. The categorisation can again be a useful framework to ascertain the categories of all the official languages, determine what tools or resources are required for each language and track the progression of the languages from the low level to higher level of resourcedness, and whether this has achieved parity among all of them.

The analysis provided for IsiNdebele presents a complexity in the categorisation of languages as any of VLR …VHR languages. The absence of a centrally driven language planning initiative funded by government and clearly linked to other areas of social, economic and scientific progress associated with industry, research institutions, and practitioners, makes it onerous to access, track and scientifically locate the development of (mostly) African languages. As a result of this lack of a centrally driven initiative, an eclectic approach to resource planning and development rests on the shoulders of independent researchers, funding agencies and non-governmental organisations providing resources for limited and often small-scale implementation initiatives. The analysis provided has taken into account all of these variables. In a very practical way this categorisation can be used (in the context of South Africa) by the Ministerial Advisory Panel that oversees the implementation of the language policy framework for higher education institutions in assessing the gaps that exist in the system and making informed decisions on where to channel the otherwise very limited funding resources for greater strategic impact.

## 5. Conclusions

The paper proposed a list of 11 features, with appropriate components of sub-scales, that capture a broader context of how a language is situated within the process of digital development of HLTs. From that, a feature-based classification of resourcedness of languages was shaped in a 5-point scale, from very low to very high-resourced. Unlike a quantitative counting of number of speakers and number of corpora and tools, it positions the resourcedness in a broader context that a language is positioned in, for these factors are intertwined with NLP resource development.

The broader context of a language affects development of human language technologies in numerous ways and its characterisation may contribute to language policy development and implementation, as well as assessment of extant policies on language technology development. Further research might investigate, inter alia, the types of resources that are available for each of South Africa's official languages and what low-hanging technology and policy development can be targeted for each according to the available resources to strategise to obtain the most realistic and effective output when augmenting the limited resources.

**Acknowledgements**

The authors would like to thank Isabelle Zaugg for feedback on an earlier draft. This work is based on the research supported in part by the National Research Foundation of South Africa (Grant Numbers 120852 and CPRR23040389063).

## Appendix: Considerations on terminology

Words and, more importantly, their semantics, matter. While the terms under-resourced, resource-scarce, less-resourced, low-resourced, or low resource all seem to be used interchangeably both in the NLP literature and in areas where NLP tools are applied, as mentioned in the Introduction, we argue that they are not. Let us analyse each one briefly, and then compare.

The notion of ‘low’ is an absolute, albeit fuzzy, measurement value. In the compound term *low resource* language, the ‘low resource’ is a property (attribute) of the language and inherently so, which is assumed to hold in the past, present, and future. This then also entails that it is something with an independent definition; e.g., one could take ‘low resource’ as the measuring stick and define ‘non-low’ (e.g., medium or high) as, say, having 10 times as much as whatever the amount of ‘low’ means, not unlike cold, temperate, and hot climates. In the compound term *low-resourced* language, the ‘resourced’ denotes a state that a certain language is in regarding its resourcedness, rather than it being an inherent immutable property of it, and states may change. That is, a language that used to be low-resourced may not be so now thanks to additional efforts; e.g., Arabic. It holds for the other end of the scale, highly resourced, as well, which could become medium resourced at a later point in time; e.g., software may not be maintained and become obsolete and corpora may be paywalled, taken offline, or not formatted into new standards. This entails that there are factors *external* to the language itself that make that language be low-resourced or high-resourced at a particular point in time.

The notion of ‘under’ is relative to something else, such as physically being ‘under’ something or abstractly like in ‘under’-performing as performing below a benchmark however that is specified. While the former use of ‘under’ may not be negative, the latter is distinctly negative, as are numerous instances of figurative speech referring to the former use, such as in ‘drinking oneself under the table’ and ‘under the thumb’. In the compound term *under-resourced* language, the ‘under’ adds a negative value connotation to the level of resourcedness as well, as it being a less than something else and not good enough compared to that something, and an entity to shift the blame onto if a task does not work out satisfactorily. What that something is exactly in terms of language resources, is typically left unspecified in the literature. Under-resourced, like low-resourced, is also a state the language is in and may get out of.

Last, the term *resource-scarce* languages, with ‘scarce’ meaning ‘insufficient for the demand’. This term also indicates that there is a specific external entity. Not a benchmark as with under-resourced, but an agent requesting it, and, conversely: if there is no demand, there is no scarcity. Since computational resources are not sought for all languages, it suggests that with the scarcity adjective, many languages would remain unclassified.

Analogous argumentation can be made at the other end of the spectrum. Highly resourced is an absolute measure; well-resourced adds a value judgement to that. The opposite of under-resourced, i.e., *over-resourced*, does, to the best of our knowledge, appear nowhere in the context of language resourcedness.

How may these observations affect use? The notion of resourcedness being an inherent property of the class of natural languages has to be discarded. A property can only be of an entity type or class, which holds for all its instances; however, with 'low resource languages', some languages do and some do not have that property, either there are subclasses of natural languages, or it is not an inherent property of natural languages. As counterexamples demonstrating change: even a dead or moribund spoken and sparsely written language can be revitalised, with Hebrew as example, and resources can be developed for newly created languages, with Esperanto as an example. Or: matters *can* change for a language.

This leaves low-resourced and under-resourced. Both can be applicable in different situations, and unlike with resource-scarce, still allow for a full classification. A language may be under-resourced *for a particular task at that time*, yet not be under-resourced for another task at the same time, i.e., it is relative to the benchmark constituted of the input requirements for that particular computational task. By elimination, in a neutral setting, *low-resourced* is then the appropriate compound term to use, which we shall use henceforth.